\def\year{2020}\relax
\documentclass[letterpaper]{article} 
\usepackage{aaai20}  
\usepackage{times}  
\usepackage{helvet} 
\usepackage{courier}  
\usepackage[hyphens]{url}  
\usepackage{graphicx} 
\usepackage{graphicx}  
\DeclareMathSizes{10}{9}{7}{5}
\usepackage{bm}
\usepackage{enumitem}
\usepackage{subfig}
\usepackage{nameref}
\usepackage{mathtools}
\usepackage{autonum}
\usepackage{tabularx}
\usepackage[flushleft]{threeparttable}
\newcolumntype{R}{>{\raggedleft\arraybackslash}X}

\newcommand*{\sov}{\operatornamewithlimits{SOV}}
\newcommand*{\pppa}{\operatornamewithlimits{PPPA}}
\newcommand*{\auroc}{\operatornamewithlimits{AUC}}
\newcommand*{\minov}{\operatornamewithlimits{minov}}
\newcommand*{\maxov}{\operatornamewithlimits{maxov}}
\newcommand*{\len}{\operatornamewithlimits{len}}
\newcommand*{\ef}{\operatornamewithlimits{F1}}

\DeclareMathOperator*{\argmax}{argmax}

\title{CAWA: An Attention-Network for Credit Attribution}
\author{Saurav Manchanda and George Karypis\\ 
University of Minnesota, Twin Cities, USA\\
\{manch043, karypis\}@umn.edu 
}

\begin{document}
\maketitle

\insert\footins{\noindent\footnotesize This is an extended version of our paper, to appear in the 34th AAAI Conference on Artificial Intelligence
(AAAI 2020).}
\begin{abstract}
Credit attribution is the task of associating individual parts in a document with their most appropriate class labels. It is an important task with applications to information retrieval and text summarization. When labeled training data is available, traditional approaches for sequence tagging can be used for credit attribution. However, generating such labeled datasets is expensive and time-consuming.
In this paper, we present \emph{Credit Attribution With Attention (CAWA)}, a neural-network-based approach, that instead of using sentence-level labeled data, uses the set of class labels that are associated with an entire document as a source of distant-supervision. CAWA combines an attention mechanism with a multilabel classifier into an end-to-end learning framework to perform credit attribution. CAWA labels the individual sentences from the input document using the resultant attention-weights. CAWA improves upon the state-of-the-art credit attribution approach by not constraining a sentence to belong to just one class, but modeling each sentence as a distribution over all classes, leading to better modeling of semantically-similar classes. Experiments on the credit attribution task on a variety of datasets show that the sentence class labels generated by CAWA outperform the competing approaches. Additionally, on the multilabel text classification task, CAWA performs better than the competing credit attribution approaches\footnote{Our code and data are available at https://github.com/gurdaspuriya/cawa.}. 
\end{abstract}
\section{Introduction}

\begin{figure}[!t]
\centering
\includegraphics[width=0.9\columnwidth]{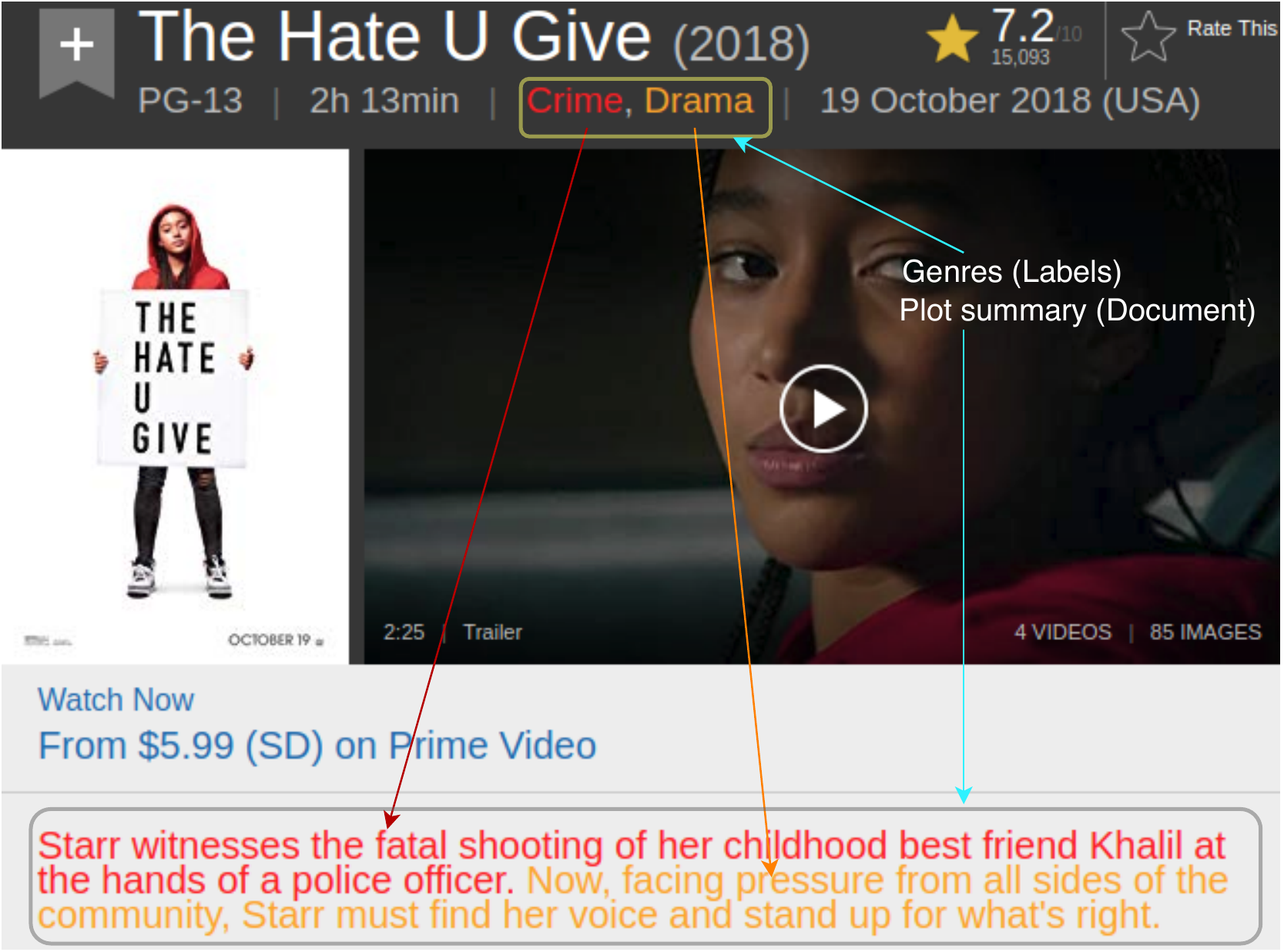}
\caption{Plot summary of the movie \textit{The Hate U Give} from the IMDB, belonging to the crime/drama genre. Red text depicts crime and orange text depicts drama.}
\label{motivation_movie}
\end{figure}

A document can be considered as a union of segments (text-pieces), where each segment tends to talk about a single topic (class). In multilabel documents, each of the segments can be associated with one or more of the document's classes. Credit attribution~\cite{ramage2009labeled} refers to the task of associating these individual segments in a document with their most appropriate class labels. For example, Figure \ref{motivation_movie} shows the screenshot of the IMDB page of the 2018 movie \textit{The Hate U Give}\footnote{https://www.imdb.com/title/tt5580266/}. The movie belongs to the crime and drama genres. As shown, each sentence in the plot summary can be mapped individually to the crime genre and drama genre. Credit attribution finds its application in many natural language processing and information retrieval tasks~\cite{hearst1997texttiling}: it can improve information retrieval (by indexing documents more precisely or by giving the specific part of a document in response to a query); and text summarization (by including information from each of the document's topics).

A straightforward way to solve credit attribution is to formulate it as a text-segment-level classification problem and collect the corresponding labeled datasets. However, manually annotating these segments (words, sentences, paragraphs, etc.) with the corresponding class labels is a tedious and expensive task. In order to reduce the need for such labeling, many methods have been developed that work in a distant-supervised fashion, such as Labeled Latent Dirichlet Allocation (LLDA)~\cite{ramage2009labeled}, Partially Labeled Dirichlet Allocation (PLDA)~\cite{ramage2011partially} Multi-Label Topic Model (MLTM)~\cite{soleimani2017semisupervised}, SEG-NOISY and SEG-REFINE~\cite{manchanda2018text}. Among them, the current state-of-the-arts are the dynamic programming based approaches SEG-NOISY and SEG-REFINE, that penalize the number of topic-switches, therefore constraining neighboring sentences to belong to the same topic. However, these approaches cannot model case where sentences can belong to multiple classes; thus, they cannot correctly model semantically similar classes.

To deal with this limitation, we developed \emph{Credit Attribution With Attention (CAWA)}, a neural-network based approach that models multi-topic segments. 
CAWA uses the class labels of a multilabel document as the source of distant-supervision, to assign class labels to the individual sentences of the document. CAWA leverages the attention mechanism to compute weights that establish the relevance of a sentence for each of the classes. The attention weights, which can be interpreted as a probability distribution over the classes, allows CAWA to capture the semantically similar classes by modeling each sentence as a distribution over the classes, instead of mapping to just one class. In addition, CAWA leverages a simple average pooling layer to constrain the neighboring sentences to belong to the same class by smoothing their class distributions. CAWA uses an end-to-end learning framework to combine the attention mechanism with the multilabel classifier.  

We evaluate the performance of CAWA on five datasets that were derived from different domains. On the credit attribution task, CAWA performs better than both MLTM and SEG-REFINE with respect to the sentence-labeling accuracy, with an average performance gain of $6.2\%$ and $9.8\%$ compared to MLTM and SEG-REFINE, respectively. On the multilabel classification task, CAWA also performs better than MLTM and SEG-REFINE. Its performance with respect to the F1 score between the predicted and the actual classes is on an average $4.1\%$ and $1.6\%$ better than MLTM and SEG-REFINE, respectively.


\section{Related Work}\label{literature}
Various unsupervised, supervised and distant-supervised methods have been developed to deal with the \textit{credit attribution} problem. The unsupervised methods do not rely on labeled training data and use approaches such as clustering or graph search to put boundaries in the locations where the transition from a topic to another happens, but no assumption is made regarding the class labels of the segments. The supervised methods rely on labeled training data, and the problem in this case can be mapped to a classification problem, the task being to predict if a sentence corresponds to beginning (or end) of a segment. The distant-supervised methods do not rely on explicitly segment-level labeled data, but use the bag-of-labels associated with a document to associate individual words/sentences in a document with their most appropriate topic labels. In  the  rest  of  this  section, we review these prior approaches for text segmentation and discuss their limitations.

Popular examples of the unsupervised approaches include TextTiling~\cite{hearst1997texttiling}, C99~\cite{choi2000advances} and GraphSeg~\cite{glavavs2016unsupervised}. TextTiling assumes that a set of lexical items are in use during the course of a topic discussion, and detects change in topic by means of change in vocabulary. Given a sequence of textual units, TextTiling computes the similarity between each pair of consecutive textual units, and locates the segment boundaries based on the minimas in the resultant similarity sequence. C99~\cite{choi2000advances} makes the same assumption, and computes the similarity between a pair of sentences based on their constituent words. These similarity values are used build a similarity matrix. Each cell in this similarity matrix is assigned a rank based on the similarity values of its neighboring cells, such that two neighboring sentences belonging to the same segment are expected to have higher ranks. C99 then uses divisive clustering on this ranking matrix to locate the boundary locations of the segments. GraphSeg~\cite{glavavs2016unsupervised} builds a semantic relatedness graph in which nodes denote the sentences and edges are created for pairs of semantically related sentences. The segments are determined by finding maximal cliques of this semantic relatedness graph.

Supervised approaches for text classification include the ones using decision trees~\cite{grosz1992some}, multiple regression analysis~\cite{hajime1998text}, exponential model~\cite{beeferman1999statistical}, probabilistic modeling~\cite{tur2001integrating} and more recently, deep neural network based approaches~\cite{badjatiya2018attention,koshorek2018text}.
The decision-tree based method uses the acoustic-prosodic features to predict the discourse structure in the text. The multiple regression analysis method~\cite{hajime1998text} uses various surface linguistic cues as features to predict the segment boundaries. The exponential model~\cite{beeferman1999statistical} uses feature selection to identify features which are used to predict segment boundaries. The selected features are lexical (including a topicality measure and a number of cue-word features). The probabilistic model~\cite{tur2001integrating} combines both lexical and prosodic cues using Hidden Markov Models and decision trees to predict the segment boundaries. The LSTM-based neural model~\cite{koshorek2018text} is composed of a hierarchy of two LSTM networks. The lower-level sub-network generates sentence representations and the higher-level sub-network is the segmentation prediction network. The outputs of the higher-level sub-network are passed on to a fully-connected network, which predicts a binary-label for each sentence, predicting whether a sentence ends a segment. The attention-based model~\cite{badjatiya2018attention} uses an LSTM with attention where the sentence representations are estimated with CNNs and the segments are predicted based on contextual information. Finally, a fully-connected network outputs a binary-label for each sentence, predicting whether a sentence begins a segment.

The methods proposed in this paper belong to the broad category of distant-supervised methods for text segmentation. Our methods use the set of labels that are associated with a document as a source of supervision, instead of using explicit segment-level ground truth information. Prior approaches proposed for distant-supervised text segmentation include Labeled Latent Dirichlet Allocation (LLDA)~\cite{ramage2009labeled}, Partially Labeled Dirichlet Allocation (PLDA)~\cite{ramage2011partially} Multi-Label Topic Model (MLTM)~\cite{soleimani2017semisupervised}, SEG-NOISY and SEG-REFINE~\cite{manchanda2018text}. We review these prior approaches below.

Labeled Latent Dirichlet Allocation (LLDA)~\cite{ramage2009labeled} is a probabilistic graphical model for \textit{credit attribution}. It assumes a one-to-one mapping between the class labels and the topics. Like Latent Dirichlet Allocation, LLDA models each document as a mixture of underlying topics and generates each word from one topic. Unlike LDA, LLDA incorporates supervision by simply constraining the topic model to use only those topics that correspond to a document’s (observed) label set. LLDA assigns each word in a document to one of the document’s labels.

Partially Labeled Dirichlet Allocation (PLDA)~\cite{ramage2011partially} is an extension of the LLDA that allows more than one topic for every class label, and some general topics that are not associated with any class. 

Multi-Label Topic Model (MLTM)~\cite{soleimani2017semisupervised} improves upon PLDA by allowing each topic to belong to multiple, one, or even zero classes probabilistically. MLTM also assigns a label to each sentence, based on the assigned topics of the constituent words. The labels of the documents are generated from the labels of its sentences. 

A common problem with the above-mentioned approaches is that they model the document as a bag of word/sentences and do not take into consideration the structure within a document, i.e., neighboring sentences tend to talk about the same topic. Recently, we proposed dynamic programming based approaches SEG-NOISY and SEG-REFINE~\cite{manchanda2018text} to segment the documents, that penalizes the number of segments, therefore constraining neighboring sentences to belong to the same topic. However, SEG-NOISY and SEG-REFINE approaches model each sentence as belonging to a single class, thus facing the problem of correctly modeling the semantically similar classes, in which case, each sentence can belong to multiple classes.

Another line of work related to the problem addressed in this paper is Rationalizing Neural Predictions~\cite{lei2016rationalizing}. The previous work selects a subset of the words in a document as a rationale for the predictions made by a neural network, where a rationale must be short. As such, this assumption makes sense for some domains, such as sentiment analysis, as a few words are sufficient to describe the sentiment. However, in our work, we do not make this assumption and develop methods for any general multilabel document. Another similar work for distant-supervised sentiment analysis~\cite{angelidis2018multiple} uses attention-mechanism to identifying positive and negative text snippets, only using the overall sentiment rating as supervision. As we explain later, in the case of multilabel documents (such as multi-aspect ratings), the vanilla-attention mechanism can assign high attention-weights to the sentences that provide negative evidence for a class. The proposed approach addresses this limitation and is well-suited for credit-attribution in multilabel documents.
\section{Definitions and Notations} \label{definitions}

\begin{table}[!t]
\small
\centering
  \caption{Notation used throughout the paper.}
  \begin{tabularx}{\columnwidth}{lX}
    \hline
Symbol   & Description \\ \hline
$D$    & Collection of multilabel documents \\
$C$    & Set of classes associated with $D$ \\
$d$    & A document from the collection $D$ \\
$|d|$    & Number of sentences in the document $d$ \\
$\mathbf{k}^w(x)$    & Key vector for the word $x$ \\
$\mathbf{v}^w(x)$    & Value vector for the word $x$ \\
$\mathbf{k}^d(i)$    & Key vector for the $i$th sentence in the document $d$ \\
$\mathbf{v}^d(i)$    & Value vector for the $i$th sentence in the document $d$ \\
$\mathbf{r}^d(c)$ &   Class-specific representation of the document $d$ for the class $c$ \\
$d[i]$    & $i$th sentence of the document $d$ \\
$y(d,c)$    & Binary indicator if the class $c$ is present in the document $d$\\
$s(d, c)$    & Probability of the document $d$ belonging to the class $c$ \\
$a(d, i, c)$ &  Attention-weight for of the $i$th sentence of $d$ for the class $c$. \\
$l(d, i)$ &  Predicted class for the $i$th sentence of the document $d$. \\
$L_d$    & Set of class labels from which each sentence in $S_d$ needs to be annotated\\
$L_C(D)$ &   Binary cross-entropy loss for predicting the classes of the documents in the collection $D$.\\
$w_c$ &   Class-specific weight for the class $c$ towards calculating $L_C(D)$.\\
$L_S(D)$ &   Attention loss, which penalizes the attention on the absent classes in a document.\\
$\alpha$ & Hyperparameter to control the relative contribution of $L_C(D)$ and $L_S(D)$ towards the final loss. \\
$\beta$ & Hyperparameter to control the relative contribution of attention-weights and document's classification probability towards sentence's classification score. \\
\hline
\end{tabularx}
  \label{tab:notation}
\end{table}
Let $C$ be a set of classes and $D$ be a set of multilabel documents. For each document $d\in D$, let $L_d \subseteq C$ be its set of classes and let $|d|$ be the number of sentences that it contains. The approach developed in this paper assumes that in multilabel documents, each sentence can be labeled with class label(s) from that document. In particular, given a document $d$ we assume that each sentence $d[i]$ can be labeled with a class $y(d,i) \in L_d$. We seek to find these sentence-level class labels, the training data being the multilabel documents and their class labels, i.e., we do not have access to the sentence-level class labels for training. 
Table \ref{tab:notation} provides a reference for the notation used throughout the paper.

\section{Credit Attribution With Attention (CAWA)}\label{proposed}

As discussed earlier, the existing approaches for credit attribution suffer from the limitations of either not modeling semantically similar classes or not exploiting the local structure within the documents. In order to address these limitations, we present a neural-network based approach \emph{Credit Attribution With Attention (CAWA)}. CAWA addresses these limitations by (i) capturing the semantically similar classes by modeling each sentence as a distribution over the classes, instead of mapping to just one class; and (ii) leveraging a simple average pooling layer to constrain the neighboring sentences to have similar class distribution; thus, leveraging the local structure within the documents.

\begin{figure*}[t]
\centerline{\includegraphics[width=0.8\textwidth]{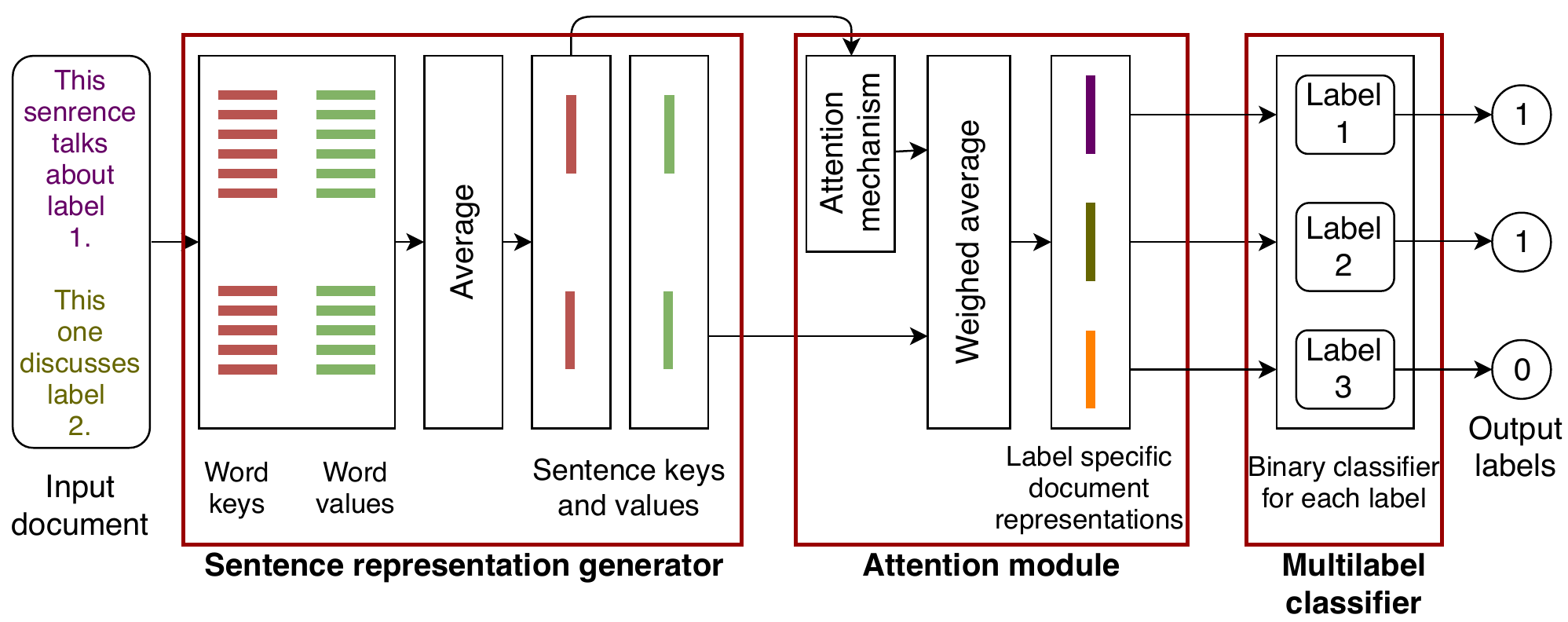}}
\caption{CAWA architecture with an example. The input document consists of two sentences, having class labels 1 and 2 respectively. The \emph{Sentence-representation generator} generates the \emph{key} and \emph{value} representation for these sentences. The attention module generates class-specific document representations using the key and value representations of the sentences. Finally, the multilabel classifier, uses these class-specific representations, to predict the correct classes of the document. Although we don't have direct supervision about the sentence-level class labels, the attention mechanism allows us to find how much each sentence is relevant to a class, that can be used to predict the sentence-level class labels.}
\label{fig:seat}
\end{figure*}
\begin{figure}[!t]
\centerline{\includegraphics[width=0.7\columnwidth]{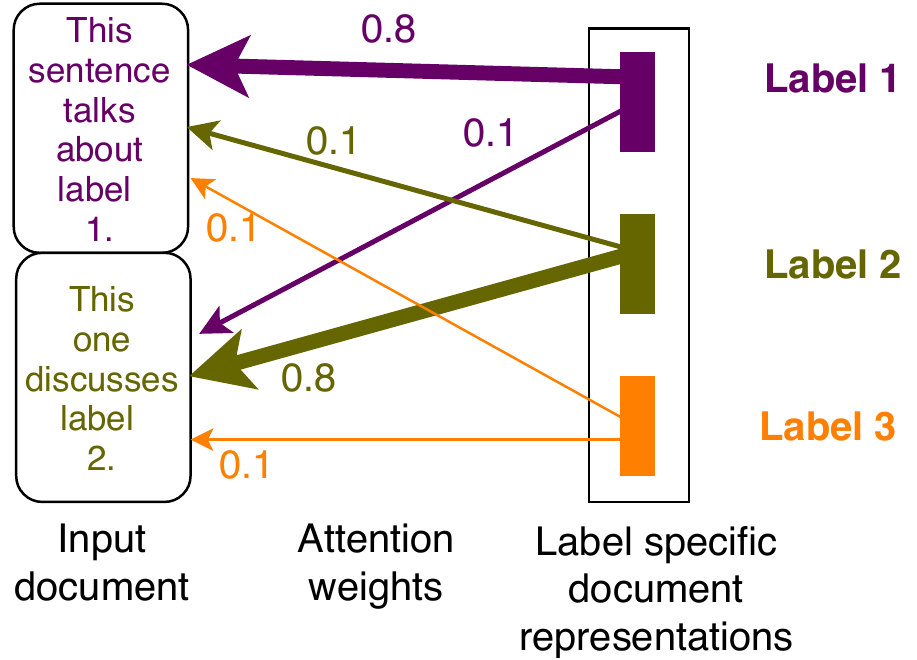}}
\caption{Example of the attention weights. Each sentence contributes towards the class-specific document representation, the extent of contribution, and hence to relevance for a class, is decided by the attention weights.}
\label{fig:attention_example}
\end{figure}

To this end, CAWA combines an attention mechanism with a multilabel classifier and uses this multilabel classifier to predict the classes of an input document. For each predicted class, the attention mechanism allows CAWA to precisely identify the sentences of the input document which are relevant towards predicting that class. Using these relevant sentences, CAWA estimates a class-specific document representation for each class. Finally, each sentence is assigned the class, for which it is most relevant, i.e., has the highest attention weight.
Figure~\ref{fig:attention_example} shown an example of sentence-labeling using the attention weights. Additionally, CAWA uses a simple average pooling layer to constrain the neighboring sentences to have similar attention weights (class distribution). 
We explain CAWA in detail in this section.

\subsection{Architecture}

CAWA consists of three components: (i) a sentence representation generator, which is responsible for generating a representation of the sentences in the input document; (ii) an attention module, which is responsible for generating a class-specific document representation from the sentence representations, and (iii) a multilabel classifier, which is responsible for predicting the classes of the document using the class-specific document representations as input. These three components form an end-to-end learning framework as shown in Figure~\ref{fig:seat}. We explain each of these components in detail in this section.

\subsubsection{Sentence-representation generator (SRG):}
The SRG takes the document as an input and generates two different representations for each sentence in the document. The two representations correspond to the \emph{keys} and the \emph{values} that will be taken as input by the attention mechanism, as explained in next section. For both \emph{keys} and \emph{values}, SRG generates the representation of a sentence as the average of the representations of the constituent words of the sentence, i.e.,

\begin{equation}
 \mathbf{k}^d(i) = \frac{1}{|d[i]|}\sum_{x\in d[i]} \mathbf{k}^w(x); \quad  \mathbf{v}^d(i) = \frac{1}{|d[i]|}\sum_{x\in d[i]} \mathbf{v}^w(x),
\end{equation}
where $d[i]$ is the $i$th sentence of document $d$, $\mathbf{k}^d(i)$ is the key-representation of $d[i]$, $\mathbf{k}^w(x)$ is the key-representation of word $x$, $\mathbf{v}^d(i)$ is the value-representation of $d[i]$ and $\mathbf{v}^w(x)$ is the value-representation of word $x$. These representations for the words are estimated during the training.

\subsubsection{Attention module:}\label{sec:attention}

\begin{figure}[!t]
\centerline{\includegraphics[width=1.0\columnwidth]{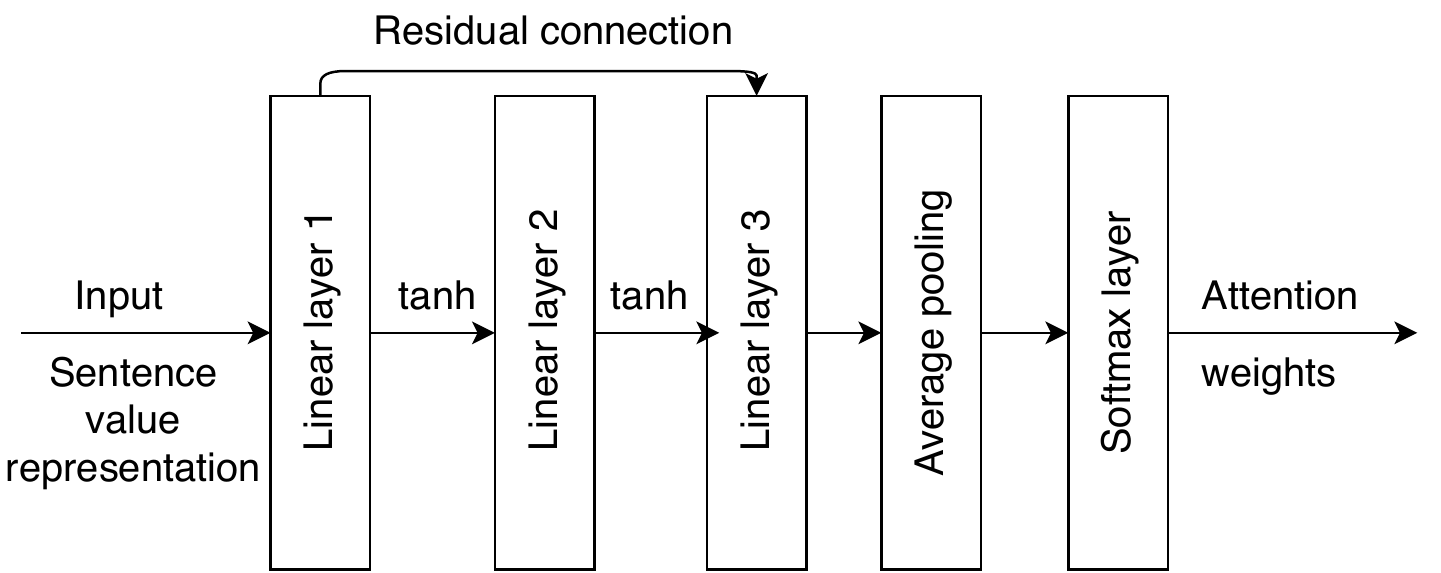}}
\caption{Architecture of the attention mechanism.}
\label{fig:attention_architecture}
\end{figure}

The attention module takes the sentence representations (\emph{keys} and \emph{values}) as input and outputs the class-specific representation of the document, one document-representation for each class. Since the different sentences have difference relevance for each class, we estimate the class-specific representations as a weighted average of the value-representations of the sentences. We calculate the \emph{attention} weights for this weighted average using a feed-forward network. Specifically, we estimate the class-specific representations as, 

\begin{equation}
 \mathbf{r}^d(c) = \sum_{i=1}^{|d|} a(d, i, c)\times\mathbf{v}^d(i),
\end{equation}
where $\mathbf{r}^d(c)$ is the class-specific representation of document $d$ for class $c$, $a(d, i, c)$ is the attention-weight for of the $i$th sentence of $d$ for class $c$.
 The feed-forward network to calculate the attention weights takes as input the \emph{key} representation of a sentence and outputs the attention weight of the sentence for each class. This feed-forward network plays the role of the sentence classifier and outputs the probability of the input sentence belonging to each of the classes, on its output layer. We implement this feed-forward network with two hidden-layers, and we use softmax on the output layer to calculate the attention-weights. To leverage the local structure within the document, i.e., to constrain the neighboring sentences to have similar class distributions, we apply average pooling before the softmax layer. Average pooling smooths out the neighboring class distributions and cancels the effect due to random variation. Note that, we can also use more flexible sequence modeling approaches, such as Recurrent Neural Networks (RNNs) to leverage the local structure. But, we choose to use a simple average pooling layer, due to its simplicity. We also add a residual connection between the first hidden layer and the output layer, which eases the optimization of the model~\cite{he2016deep}. The architecture for the attention mechanism is shown in Figure~\ref{fig:attention_architecture}.

\subsubsection{Multilabel classifier:}

\begin{figure}[!t]
\centerline{\includegraphics[width=1.0\columnwidth]{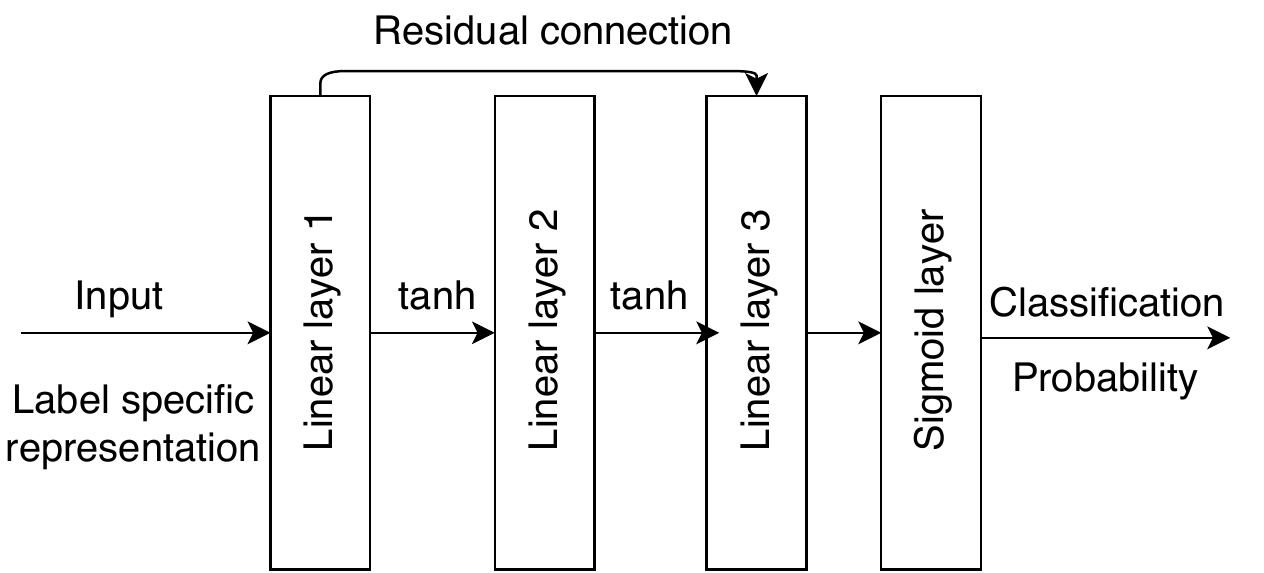}}
\caption{Architecture of the per-class binary classifier.}
\label{fig:classification_architecture}
\end{figure}
Several architectures and loss-functions have been proposed for multilabel classification, such as Backpropagation for Multi-Label Learning (BP-MLL)~\cite{zhang2007ml}. However, as the focus of this paper is credit attribution and not multilabel classification, we simply implement the multilabel classifier as a separate binary classifier for each class. Therefore, each binary classifier predicts whether a particular class is present in the document or not. The input to each of these binary classifiers is the class-specific representation, which is the output of the attention module. We implement each of these binary classifiers as a feed-forward network with two hidden layers and use sigmoid on the output layer to predict the probability of the document belonging to that class. The architecture for the class-specific binary classifiers is shown in Figure~\ref{fig:classification_architecture}.

\subsection{Model estimation}
To quantify the loss for predicting the classes of a document, we minimize the weighted binary cross-entropy loss~\cite{nam2014large}, which is a widely used loss function for multilabel classification. The weighted binary cross-entropy loss associated with all the documents in collection $D$ is given by:
\begin{multline}
L_C(D) = -\frac{1}{|D|}\sum_{d\in D}\sum_{c\in C} w_c( y(d, c) \log(s(d, c))\\ + (1 - y(d, c))\log(1 - s(d, c))),
\end{multline}
where $y(d, c) = 1$ if the class $c$ is present in document $d$, and $y(d, c) = 0$ otherwise, $s(d, c)$ is the prediction probability of document $d$ belonging to class $c$, and $w_c$ is the class-specific weight for class $c$. This weight $w_c$ is used to handle the class imbalance by increasing the importance of infrequent classes (upsampling), and we empirically set it to $w_c = \sqrt{|D|/n_c}$, where $n_c$ is the number of documents belonging to the class $c$. Note that we require the sentences to be labeled with the class which they describe. However, the attention mechanism can also assign high attention-weights to the sentences that provide a negative-signal for a class. For example, if a document can exclusively belong to only one class A and B, the text describing one of the classes (say A) will also provide a negative signal for the other class (B), and hence, will get high attention-weight for both classes A and B. To constraint that the attention is only focused on the classes that are actually present in the document, we introduce \emph{attention loss}, which penalizes the attention on the absent classes, and is given by

\begin{equation}
L_S(D) = -\frac{1}{|D|}\sum_{d\in D}\frac{1}{|d|}\sum_{i=1}^{|d|}\sum_{c\in C} (1 - y(d, c))\log(1 - a(d, i, c)),
\end{equation}
where $a(d, i, c)$ is the attention weight for class $c$ on the $i$th sentence of the document $d$.
To estimate the CAWA, we minimize the weighted sum of both $L_C(D)$ and $L_S(D)$, given by $L(D) = \alpha L_C(D) + (1 - \alpha)L_S(D), $
where $\alpha$ is a hyperparameter to control the relative contribution of $L_C(D)$ and $L_S(D)$ towards the final loss.

\begin{table*}[!t]
\footnotesize
\centering
  \caption{Dataset statistics.}
  \begin{tabularx}{\linewidth}{lRRRRRRRRRR}
    \hline
    Dataset&Vocab size&Number of classes&Number of training documents&Average number of classes in training documents&Average number of words in training documents&Average number of sentences in training documents&Average number of words per sentence in training documents&Number of test documents (classification)&Number of test documents (credit a.)&Average number of classes in test documents (credit a.)\\
    \hline
    Movies&9,568&6&3,834&2.23&88.9&12.5&7.1&959&488&2.26\\
    Ohsumed&13,253&23&12,800&2.41&103.3&8.2&12.5&3,200&500&2.39\\
    TMC2007&10,686&22&15,131&2.69&104.7&12.3&8.5&3,783&500&2.66\\
    Patents&5,681&8&9,257&2.25&37.9&5.1&7.4&2,314&500&2.22\\
    Delicious&8,778&20&6,871&2.54&128.3&19.2&6.7&1,718&488&2.48\\
    \hline
\end{tabularx}
  \label{tab:dataset}
\end{table*}
\subsection{Segment inference}
We can directly use the estimated attention-weights to assign a class to each sentence, corresponding to the class with the maximum attention-weight. However, to ensure the consensus between the predicted sentence-level classes and document's classes, we use a linear combination of the attention-weights and document's predicted class-probabilities to assign a class to each sentence, i.e., 

\begin{equation}\label{eq:sent_score}
l(d, i) = \argmax_c (\beta\times a(d, i, c) + (1-\beta)\times y(d,c)),
\end{equation}
where $l(d, i)$ is the predicted class for the $i$th sentence of $d$ and $\beta$ is a hyperparameter to control the relative contribution of attention-weights and document's classification probability. Additionally, $y(d,c)$ acts as a global bias term, and makes the sentence-level predictions less prone to random variation in the attention weights.

\section{Experimental methodology}\label{experiments}
\subsection{Datasets}
We performed experiments on five multilabel text datasets belonging to different domains as described below:
\begin{itemize}[leftmargin=*]
    \item \textit{Movies~\cite{bamman2014learning}}: This dataset contains movie plot summaries extracted from Wikipedia and corresponding genres extracted from Freebase. We randomly take a subset of the movies from this dataset corresponding to six common genres: \emph{Romance Film}, \emph{Comedy}, \emph{Action}, \emph{Thriller}, \emph{Musical}, \emph{Science Fiction}.
    \item \textit{Ohsumed~\cite{hersh1994ohsumed}:} The Ohsumed test collection is a subset of the MEDLINE database. The labels correspond to 23 Medical Subject Headings (MeSH) categories of cardiovascular diseases group.
    \item \textit{TMC2007\footnote{https://c3.nasa.gov/dashlink/resources/138/}:} This is the dataset used for the SIAM 2007 Text Mining competition. The documents are the aviation safety reports corresponding to the one or more problems that occurred during certain flights. There are a total of 22 unique labels. 
    \item \textit{Patents\footnote{http://www.patentsview.org/download/}}: This dataset contains brief summary text of the parents and the labels correspond to the associated Cooperative Patent Classification (CPC) group labels. We randomly take a subset of summaries corresponding to the eight CPC groups: \emph{A: Human Necessities}, \emph{B: Operations and Transport}, \emph{C: Chemistry and Metallurgy}, \emph{D: Textiles}, \emph{E: Fixed Constructions}, \emph{F: Mechanical Engineering}, \emph{G: Physics}, \emph{H: Electricity}.
    \item \textit{Delicious~\cite{zubiaga2009content}:} This data set contains tagged web pages retrieved from the social bookmarking site delicious.com. Tags for the web pages in this data set are not selected from a predefined set of labels; rather, users of the website delicious.com bookmarked each page with single word tags. We randomly choose documents corresponding to 20 common tags as our class labels: \emph{humour}, \emph{computer}, \emph{money}, \emph{news}, \emph{music}, \emph{shopping}, \emph{games}, \emph{science}, \emph{history}, \emph{politics}, \emph{lifehacks}, \emph{recipes}, \emph{health}, \emph{travel}, \emph{math}, \emph{movies}, \emph{economics}, \emph{psychology}, \emph{government}, \emph{journalism}.
\end{itemize}

For both the credit attribution and multilabel classification tasks, we used the same training and test dataset split as used in~\cite{manchanda2018text}. For the multilabel classification task, both training and test data are the documents with at least two classes associated with each document. For the credit attribution, the test dataset is synthetic, and each test document corresponds to multiple single-label documents concatenated together (thus, giving us ground truth sentence  class labels for a document). Additionally, we also use a validation dataset, created in a similar manner to this test dataset, for the hyperparameter selection. 
Table \ref{tab:dataset} reports the statistics of these datasets.

\subsection{Baselines}
Although not a lot of approaches have been developed that are specifically designed to solve the credit attribution problem, any multilabel classifier can be used to perform credit attribution, by training on the multilabel documents and predicting the classes of the individual sentences. Thus, apart from the credit attribution specific approaches, we compare CAWA against several multilabel classification approaches. We selected our baselines from diverse domains such as graphical models, deep neural networks, dynamic programming as well as classical approaches for text classification such as Multinomial Naive Bayes. Specifically, we compare CAWA against the following baselines:
\begin{itemize}[leftmargin=*]
    \item \textit{SEGmentation with REFINEment (SEG-REFINE)}~\cite{manchanda2018text}: SEG-REFINE is dynamic programming based approach to segment the documents, that penalizes the number of segments, therefore constraining neighboring sentences to belong to the same topic.
    \item \textit{Multi-Label Topic Model (MLTM)}~\cite{soleimani2017semisupervised}: MLTM is a probabilistic generative approach, that generates the classes of a document from the classes of its constituent sentences, which are further generated from the classes of the constituent words. 
    \item \textit{Deep Neural Network with Attention (DNN+A):} As mentioned earlier, any multilabel classifier can be used to perform credit attribution, by training on the multilabel documents and predicting the classes of the individual sentences. Thus, we compare CAWA against a deep neural network based multilabel classifier. For a fair comparison, we use the same architecture as CAWA for DNN+A, except the components specific to CAWA (attention loss and average pooling layer).
    \item \textit{Deep Neural Network without Attention (DNN-A):} DNN-A has the same architecture as DNN+A, except the attention, i.e., each class gives equal emphasis on all the sentences.
    \item \textit{Multi-Label k-Nearest Neighbor (ML-KNN)}~\cite{zhang2007ml}: ML-KNN is a popular method for multilabel classification. It uses k nearest neighbors to a test example and uses Bayesian inference to assign classes to the text example.
    \item \textit{Binary Relevance - Multinomial Naive Bayes (BR-MNB):} Binary relevance is also a popular approach for multilabel classification,  and amounts to independently training a binary classifier for each class. The prediction output is the union of all per class classifiers. We use Multinomial Naive Bayes as the per class binary classifier, which is a popular classical approach for text classification.
\end{itemize}
\subsection{Performance Assessment Metrics}

\subsubsection{Credit attribution: }
\begin{figure}[!t]
\centerline{\fbox{\includegraphics[width=1.0\columnwidth]{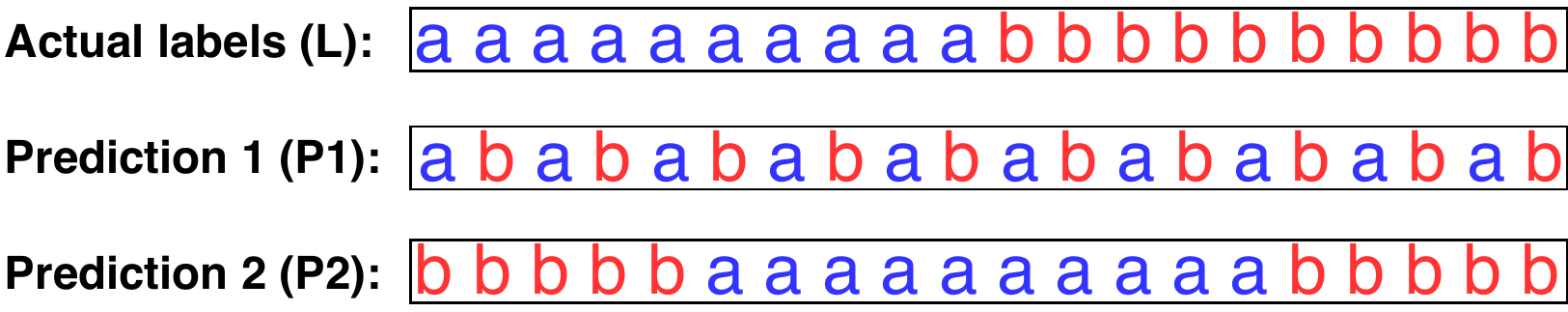}}}
\caption{Illustration of the $\sov$ and $\pppa$ metrics. Predictions P1 and P2 score the same on the $\pppa$ metric with $\pppa(L, P1) = \pppa(L, P2) = 0.5$. But, P2 being more aligned with the ground truth class labels, scores higher than P1 on the $\sov$ metric, with $\sov(L, P1) = 0.5\text{ and }\sov(L, P2) = 0.83$.}
\label{sov_illustration}
\end{figure}

For evaluation on the credit attribution task, we look into two different metrics. The first is \emph{per-point prediction accuracy} ($\pppa$) and the second is Segment OVerlap score ($\sov$)\cite{rost1994redefining}. $\pppa$ corresponds to the fraction of sentences that are predicted correctly and is defined as:
\begin{equation}
    \pppa(S_1, S_2) = \frac{1}{N}\sum_{i=1}^N 1(S_1(i) == S_2(i)).
\end{equation}
As a single-point measure, $\pppa$ does not take into account the correlation between the neighboring sentences. On the other hand, $\sov$ measures how well the observed and the predicted segments align with each other. $\sov$ is defined as
\begin{equation}
    \sov(S_1, S_2) = \frac{1}{N}\sum_{\substack{s_1\in S_1\\s_2 \in S_2\\(s_1, s_2) \in s}}\frac{\minov(s_1, s_2) + \delta(s_1, s_2)}{\maxov(s_1, s_2)}\times \len(s_1),
\end{equation}
where $N$ is the total number of sentences in the document we are segmenting, $S_1$ is the observed segmentation, and $S_2$ is the predicted segmentation. The sum is taken over all segment pairs $(s_1, s_2) \in s$ for which $s_1$ and $s_2$ overlap on at least one point. The actual overlap between the $s_1$ and $s_2$ is $\minov(s_1, s_2)$, that is, the number of points both segments have in common, while $\maxov(s_1, s_2)$ is the total extent of both segments. The accepted variation $\delta(s_1, s_2)$ brings robustness in case of minor deviations at the ends of segments. $\delta(s_1, s_2)$ is defined as~\cite{zemla1999modified}:
\begin{equation}
    \delta(s_1, s_2)=\min
    \begin{cases}
      \maxov(s_1, s_2) - \minov(s_1, s_2). \\
      \hfil \minov(s_1, s_2). \\
      \hfil \lfloor\len(s_1)/2\rfloor. \\
      \hfil \lfloor\len(s_2)/2\rfloor.
    \end{cases}
\end{equation}
Compared to the $\pppa$ metric, $\sov$ penalizes fragmented segments and favors continuity in the predictions. For example, prediction errors at the end of segments will be penalized less by $\sov$ than the prediction errors in the middle of the segments. Consequently, $\sov$ favors contiguous segments, at the cost of mislabeling individual sentences. Figure \ref{sov_illustration} illustrates the difference between the $\sov$ and $\pppa$ metrics. 

\subsubsection{Multilabel classifiation: }
To evaluate CAWA on the multilabel classification task, we looked into three metrics: $\ef$, $\auroc_{\mu}$ and $\auroc_{M}$. For a given document, $\ef$ score is the harmonic mean of the precision and recall based on the predicted classes and the observed classes. We report the mean of $\ef$ score over all the test documents. Area Under the Receiver Operating Characteristic Curve ($\auroc$)~\cite{bradley1997use} gives the probability that a randomly chosen positive example ranks above a randomly chosen negative example. We report $\auroc$ both under the micro ($\auroc_{\mu}$) and macro ($\auroc_{M}$) settings. $\auroc_{M}$ computes the metric independently for each class and then takes the average (hence treating all classes equally), whereas $\auroc_{\mu}$ aggregates the contributions of all classes to compute the metric. 

\subsection{Parameter selection}
\begin{table}[!t]
\centering
\scriptsize
  \caption{Hyperparameter values.}
  \begin{tabularx}{1.0\columnwidth}{Xrrrrr}
    \hline
 & $\alpha$ & $\beta$ & $k$ & $\alpha$ & $m$\\
Dataset   & (CAWA) & (CAWA) & (ML-KNN) & (SEG-REFINE) & (MLTM)\\ \hline
Movies    & 0.2 & 0.1 &100 &0.50 &120\\
Ohsumed   & 0.1 & 0.1 &20 &0.40 &90 \\
TMC2007   & 0.1 & 0.3 &50 &0.40 &90\\
Patents   & 0.5 & 0.3 &50 &0.45 &110\\
Delicious & 0.1 & 0.2 &20 &0.50 &70\\ \hline
\end{tabularx}
  \label{tab:hyperparameters}
\end{table}

We chose the values of $\alpha$ and $\beta$ individually for all the datasets, using grid search in the range $\{0.0,0.1,\ldots ,1.0\}$, based on the best validation $\sov$ score.
For CAWA, DNN+A, and DNN-A, the number of nodes in the each of the hidden layer, all representations' length, as well as the batch size for training the CAWA was set to 256. For regularization, we used a dropout~\cite{srivastava2014dropout} of $0.5$ between all layers, except the output layer. For optimization, we used the ADAM~\cite{kingma2014adam} optimizer. We trained all the models for 100 epochs, with the learning-rate set to 0.001. The keys and values embeddings are initialized randomly. For average pooling in CAWA, we fixed the kernel-size to three. For ML-KNN, we used cosine similarity measure to find the nearest neighbors which is a commonly used similarity measure for text documents. We chose the number of neighbors ($k$) for ML-KNN based on the best $\sov$ score of the validation set. For ML-KNN and BR-MNB, we used the implementation as provided by scikit-multilearn\footnote{http://scikit.ml/}. The hyperparamemer of MLTM is the number of topics ($m$), and the hyperparameter for SEG-REFINE is the segment creation penalty ($\alpha$). We find these hyperparameters for MLTM and SEG-REFINE based on the best validation $\sov$ score. The chosen values of hyperparameters for different datasets are shown in Table \ref{tab:hyperparameters}. 

\begin{table}[!t]
\scriptsize
\centering
  \caption{Performance comparison results.}
  \begin{threeparttable}
          \begin{tabularx}{\columnwidth}{lXrrrrr}
            \hline
        Dataset   & Model\textsuperscript{*}    & $\sov$  & $\pppa$   & $\ef$   & $\auroc_{\mu}$ & $\auroc_{M}$ \\ \hline
        Movies & CAWA & \textbf{0.50} & 0.38 & \textbf{0.65} & 0.81  & 0.78 \\ 
        & SEG-REF & 0.49 & 0.36 & 0.63 & 0.81 & 0.80 \\ 
        & MLTM     & \textbf{0.50} & \textbf{0.40} & \textbf{0.65} & 0.82 & 0.80 \\ 
        & DNN+A     & 0.33 & 0.27  & 0.62 & 0.84 & 0.82 \\  
        & DNN-A     & 0.33 & 0.27  & 0.61 & \textbf{0.85} & 0.83 \\   
        & ML-KNN     & 0.38 & 0.30  & 0.63 & 0.83 & 0.81 \\    
        & BR-MNB     & 0.39 & 0.31  & 0.53 & 0.82  & \textbf{0.84} \\  \hline
        Ohsumed& CAWA & \textbf{0.65} & \textbf{0.55}  & 0.64 & 0.93 &0.89 \\ 
        & SEG-REF & 0.63 & 0.47 &0.65 & \textbf{0.94} & \textbf{0.92} \\ 
        & MLTM     & 0.56 & 0.47 & 0.60 & 0.93 & 0.91 \\
        & DNN+A     & 0.44 & 0.37 & \textbf{0.67} & \textbf{0.94} & \textbf{0.92} \\ 
        & DNN-A     & 0.33 & 0.31 & 0.58 & \textbf{0.94} & \textbf{0.92} \\ 
        & ML-KNN     & 0.48 & 0.38 & 0.59 & 0.90 & 0.87 \\    
        & BR-MNB     & 0.29 & 0.30 & 0.31 & 0.82 & 0.71 \\  \hline
        TMC2007& CAWA & 0.56 & \textbf{0.47} & 0.68 & 0.95 & 0.91 \\
        & SEG-REF & \textbf{0.59} & 0.44 & 0.68 & 0.95 & 0.90 \\ 
        & MLTM     & 0.49 & 0.43 & 0.64 & \textbf{0.96} & \textbf{0.92} \\ 
        & DNN+A     & 0.43 & 0.37 & 0.68 & \textbf{0.96} & \textbf{0.92} \\  
        & DNN-A     & 0.35 & 0.34  & 0.59 & \textbf{0.96} & \textbf{0.92} \\    
        & ML-KNN     & 0.45 & 0.35  & \textbf{0.71} & 0.95 & 0.89 \\    
        & BR-MNB     & 0.30 & 0.33 & 0.62 & 0.89 & 0.72 \\  \hline
        Patents& CAWA & \textbf{0.58} & \textbf{0.50} & 0.61 & 0.88 & 0.86 \\ 
        & SEG-REF & 0.56 & 0.45 & 0.61 & 0.86 & 0.85 \\ 
        & MLTM     & 0.55 & 0.48     & 0.59 & 0.85 & 0.84 \\ 
        & DNN+A     & 0.53 & 0.43 & \textbf{0.64} & \textbf{0.89} & 0.87 \\  
        & DNN-A     & 0.51 & 0.42 & 0.63 & \textbf{0.89} & \textbf{0.88} \\   
        & ML-KNN     & 0.45 & 0.37 & 0.51 & 0.82 & 0.80 \\    
        & BR-MNB     & 0.50 & 0.43 & 0.50 & 0.87 & 0.86 \\ \hline  
        Delicious& CAWA & \textbf{0.50} & \textbf{0.39} &\textbf{0.52} & 0.85 & 0.84 \\ 
        & SEG-REF & 0.48 & 0.36 & 0.49 & 0.85 & 0.85 \\
        & MLTM     & 0.49 & 0.37 & 0.50 & 0.84 & 0.83 \\ 
        & DNN+A     & 0.22 & 0.18 & 0.38 & 0.87 & 0.86 \\  
        & DNN-A     & 0.21 & 0.17  & 0.36 & \textbf{0.88} & \textbf{0.87} \\    
        & ML-KNN     & 0.24 & 0.19  & 0.35 & 0.82 & 0.80 \\    
        & BR-MNB     & 0.25 & 0.19  & 0.05 & 0.76 & 0.73 \\ \hline
        \end{tabularx}
        \begin{tablenotes}
         \item[*] The models CAWA, SEG-REFINE (abbreviated SEG-REF above) and MLTM have been specifically designed to solve the credit attribution problem, while the models DNN+A, DNN-A, ML-KNN and BR-MNB are multilabel classification approaches.
        \end{tablenotes}
  \end{threeparttable}
          \label{tab:segmentation}
\end{table}

\section{Results and Discussion}\label{results}

\subsection{Credit attribution}
The metrics $\sov$ and $\pppa$ in Table \ref{tab:segmentation} show the performance for various methods on the credit attribution task. The credit attribution specific approaches (CAWA, SEG-REFINE, and MLTM) perform considerably better than the other multilabel approaches (DNN+A, DNN-A, ML-KNN, and BR-MNB). CAWA performs better than the SEG-REFINE and MLTM on the $\pppa$ metric for the \emph{Ohsumed}, \emph{TMC2007}, \emph{Patents} and \emph{Delicious} datasets. The average performance gain for the CAWA on the $\pppa$ is $6.2\%$ compared to MLTM and  $9.8\%$ compared to SEG-REFINE. Additionally, CAWA also performs at par, if not better, than the SEG-REFINE and MLTM on the $\sov$ metric. This shows that CAWA is able to find contiguous segments, without compromising on the sentence-level accuracy. 

To validate our hypotheses that CAWA can accurately model the semantically similar classes as compared to the SEG-REFINE, we looked into the performance of both CAWA and SEG-REFINE on the two most similar classes for each of the \emph{Ohsumed}, \emph{Patents} and \emph{Delicious} datasets. To measure the similarity between the two classes, we calculated the Jaccard similarity~\cite{jaccard1901etude} between these classes, based on the number of documents in which they occur. For each of these selected classes, we calculated the F1 score based on the predicted and actual classes of the sentences in the segmentation dataset. Table~\ref{tab:similar_labels} shows the results for this analysis. For the \emph{Ohsumed} dataset, the two selected classes are Nutritional/Metabolic disease and Endocrine disease, which are very similar. Likewise, the selected classes for the \emph{Patents} and \emph{Delicious} dataset are also similar. We see that, for all the selected classes, CAWA performs better than SEG-REFINE, illustrating the effectiveness of CAWA on modeling semantically similar classes. We further investigate the effect of various parameters of CAWA on the credit attribution task in the~\nameref{sec:ablation} section.

\begin{table}[!t]
\footnotesize
\centering
\caption{Sentence classification performance on similar classes.}
  \begin{tabularx}{\columnwidth}{XXlR}
    \hline
        Dataset   & Class & Model    & F1\\ \hline
        Ohsumed   & Nutritional/ & CAWA & \textbf{0.68} \\ 
                  & metabolic  & SEG-REFINE & 0.64\\ 
                  & Endocrine & CAWA & \textbf{0.39} \\ 
                  & disease   & SEG-REFINE & 0.26\\ \hline
        Patents   & Electricity & CAWA & \textbf{0.53} \\ 
                  &    & SEG-REFINE & 0.48\\ 
                  & Physics & CAWA & \textbf{0.41} \\ 
                  &    & SEG-REFINE & 0.33\\ \hline
        Delicious & Health  & CAWA & \textbf{0.50} \\ 
                  &    & SEG-REFINE & 0.47\\ 
                  & Recipes & CAWA& \textbf{0.62} \\ 
                  &    & SEG-REFINE & 0.58\\ \hline
        \end{tabularx}
  \label{tab:similar_labels}
\end{table} 

In addition, DNN+A also performs considerably better than the DNN-A on both $\sov$ and $\pppa$ metrics for all the datasets.
This shows the effectiveness of the proposed attention architecture on modeling the multilabel documents.

\subsection{Multilabel classification}

\begin{figure*}[!t]
\centerline{\includegraphics[width=1.0\textwidth]{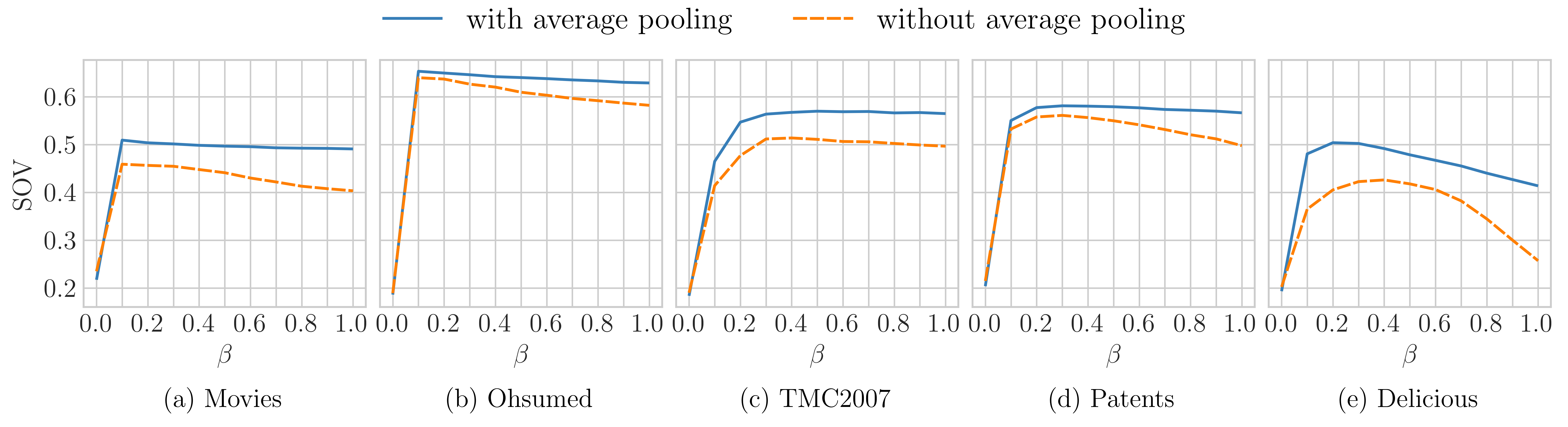}}
\caption{Change in the $\sov$ with $\beta$ as it is increased from $0$ to $1$ for the two cases (i) average-pooling layer is used, and (ii) average-pooling layer is not used. The plots correspond the the values of $\alpha$ as reported in Table~\ref{tab:hyperparameters}.}
\label{fig:beta_sov}
\end{figure*}
The metrics $\ef$, $\auroc_{\mu}$ and $\auroc_{M}$ in Table \ref{tab:segmentation} show the performance of different methods on the classification task. Similar to the credit attribution task, CAWA, in general, performs better than the competing credit attribution approaches (SEG-REFINE and MLTM) on the $\ef$ metric, with an average performance gain of $4.1\%$ over MLTM and $1.6\%$ over SEG-REFINE. This shows that the classes predicted for the sentences by CAWA correlate better with the document classes as compared to the classes predicted by the competing credit attribution approaches. This can be attributed to the way we calculate the sentence classes (Equation~\ref{eq:sent_score}), which ensures the consensus between the predicted sentence-level and document-level classes. Additionally, CAWA performs at par with the competing credit attribution approaches on the $\auroc_{\mu}$ and $\auroc_{M}$ metrics, further illustrating the effectiveness of CAWA.

Compared to the approaches specific to the multilabel classification task, either CAWA or DNN+A achieve the best performance on the $\ef$ metric on all but the TMC2007 dataset, where ML-KNN achieves the best performance.  This further verifies the effectiveness of the proposed attention architecture on correctly modeling the multilabel documents. 
On the $\auroc$ metrics, we see that DNN+A outperforms CAWA. This is the result of attention loss, which while helping the network to perform credit attribution, damages its ability to perform global document classification. As discussed earlier, the vanilla attention mechanism (as used in the DNN+A) can assign high attention-weights to the sentences that provide a negative-signal for a class. Attention loss constrains that the attention is only focused on the classes that are actually present in the document. Thus, while DNN+A also leverages the negative signals for the classes to make its predictions, CAWA, by design, ignores these negative signals, which adversely affects its multi-label classification performance. We further investigate the effect of attention loss on the multilabel classification task in the~\nameref{sec:ablation} section.

\subsection{Ablation study}\label{sec:ablation}
\subsubsection{Effect of average pooling and $\beta$:}

\begin{figure*}[!t]
\centerline{\includegraphics[width=1.0\textwidth]{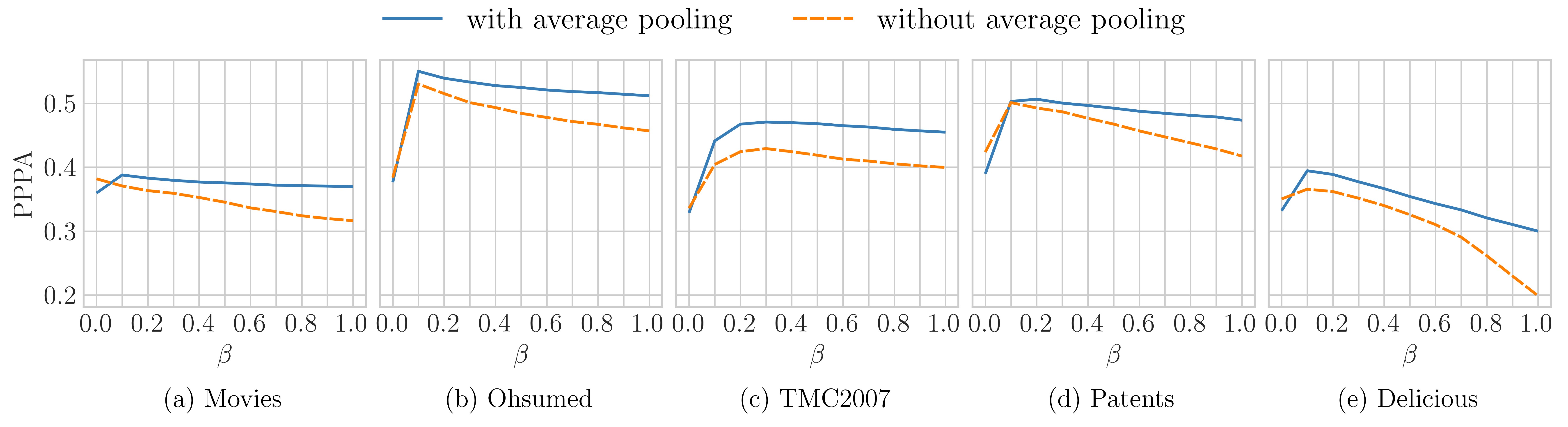}}
\caption{Change in the $\pppa$ with $\beta$ as it is increased from $0$ to $1$ for the two cases (i) average-pooling layer is used, and (ii) average-pooling layer is not used. The plots correspond the the values of $\alpha$ as reported in Table~\ref{tab:hyperparameters}.}
\label{fig:beta_pppa}
\end{figure*}
Figure~\ref{fig:beta_sov} shows the change in $\sov$ metric with change in $\beta$ for all the datasets. For each dataset, we plot the $\sov$ metric as $\beta$ is increased from $0.0$ to $1.0$ for the two cases (i) average-pooling layer is used, and (ii) average-pooling layer is not used. For both the cases, when $\beta=0$, each sentence gets the same class, which is the class with the maximum prediction probability for the complete document. As $\beta$ increases, the effect of the attention-weights starts pitching in, leading to each sentence getting its own class, thus a sharp jump in the performance on the $\sov$ metric.  However, as the $\beta$ increases, the contribution of attention weights outpowers the overall document class probabilities, and the predicted sentence-classes become more prone to noise in the attention weights, thus leading to performance degradation for large $\beta$. 

Comparing the performance curves of the case when the average-pooling layer is used to the one when it is not used, the average pooling leads to better performance for all values of $\beta$. Thus, average pooling effectively constrains the nearby sentences to have similar attention weights, leading to better performance on the $\sov$ metric.

The similar behaviour is shown by the performance on the $\pppa$ metric, as depicted in Figure~\ref{fig:beta_pppa}.

\begin{figure}[!t]
\centerline{\includegraphics[width=1.0\columnwidth]{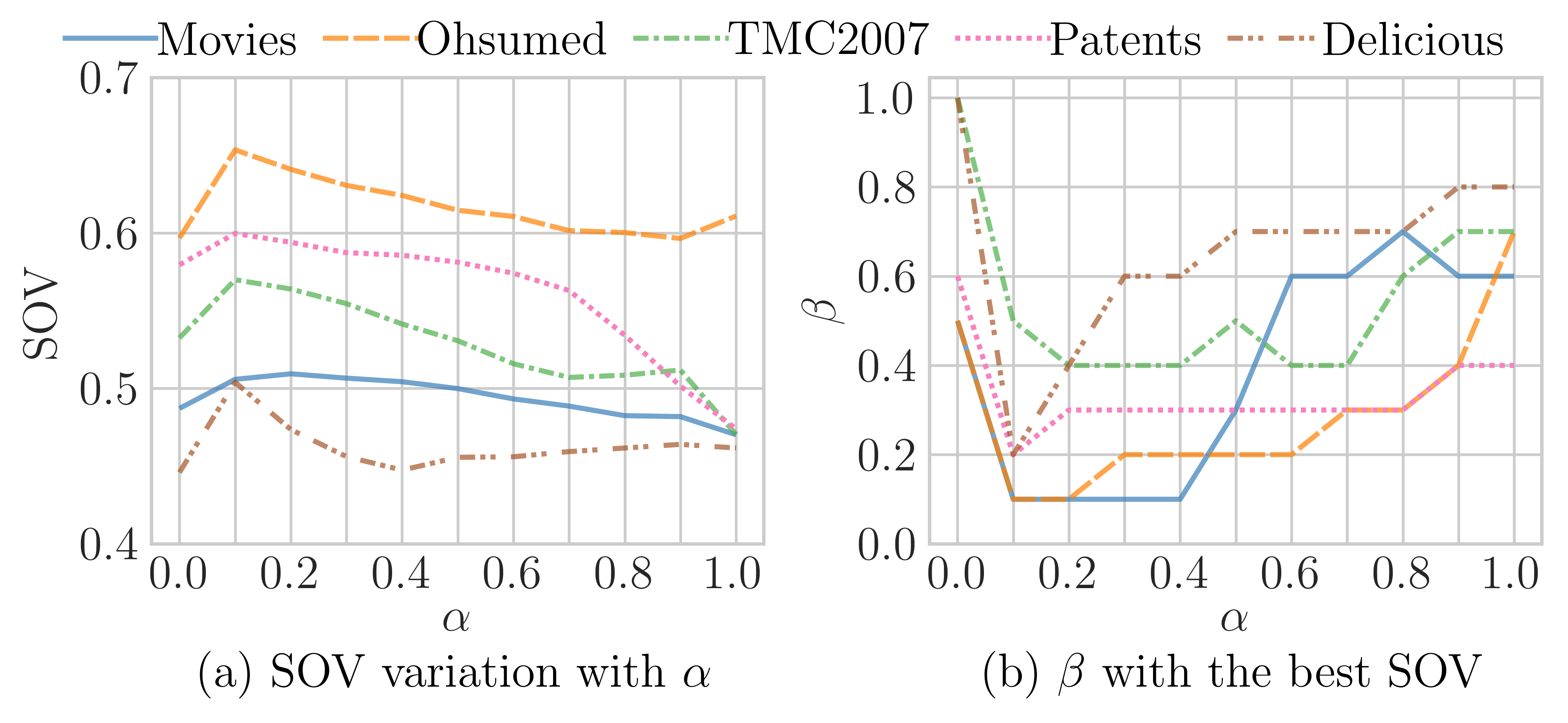}}
\caption{Sub-figure (a) shows the change in $\sov$ with change in $\alpha$. 
Sub-figure (b) shows the $\beta$ values for which the maximum $\sov$ is obtained for each $\alpha$.}
\label{fig:alpha_sov}
\end{figure}

\begin{figure}[!t]
\centerline{\includegraphics[width=1.0\columnwidth]{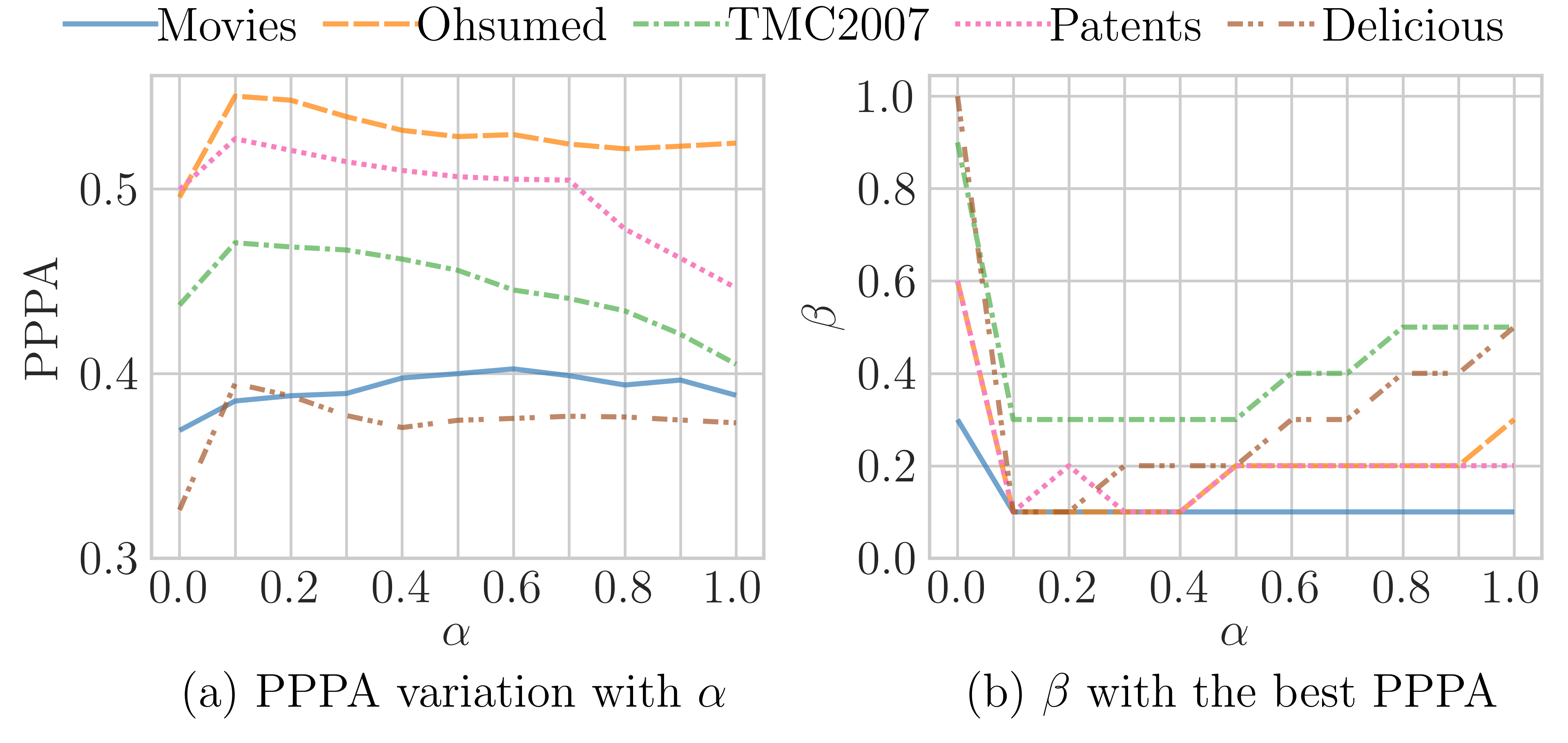}}
\caption{Sub-figure (a) shows the change in $\pppa$ with change in $\alpha$. 
Sub-figure (b) shows the $\beta$ values for which the maximum $\pppa$ is obtained for each $\alpha$.}
\label{fig:alpha_pppa}
\end{figure}

\subsubsection{Effect of $\alpha$:}

Figure~\ref{fig:alpha_sov} shows the change in performance on the $\sov$ metric with change in $\alpha$ for all the datasets. Figure~\ref{fig:alpha_sov}(a) reports the maximum value of $\sov$ for each $\alpha$ over all the $\beta$ values. Figure~\ref{fig:alpha_sov}(b) reports the corresponding value of $\beta$ for each $\alpha$ that gives the maximum performance on the $\sov$ metric. For all the cases, as the $\alpha$ increases from $0.0$ to $0.1$, $\sov$ shows a sharp increase, which can be attributed to the effect of classification loss ($L_C(D)$) pitching in. Additionally, we see that as the $\alpha$ increases, the corresponding value of $\beta$ giving the maximum performance also increases in general. As the $\alpha$ increases, the contribution of attention loss decreases, thus requiring more contribution from the attention weights to accurately predict the sentence classes. This explains the increase in the values of $\beta$ values, as the value of $\alpha$ increases. The exceptionally high value of $\beta$ when $\alpha=0$ can be explained as follows: $\alpha=0$ corresponds to the case when we are only minimizing the attention loss ($L_S(D)$), and ignoring the loss for predicting the document's classes ($L_C(D)$). The multilabel classifier does not get trained at all in this case, leading to $y(d,c)$ getting random values. Therefore, $\beta$ takes large values to ignore the contribution of $y(d,c)$ (which is random) towards the sentence-level labels, so as to make correct predictions. 

\begin{figure}[!t]
\centerline{\includegraphics[width=1.0\columnwidth]{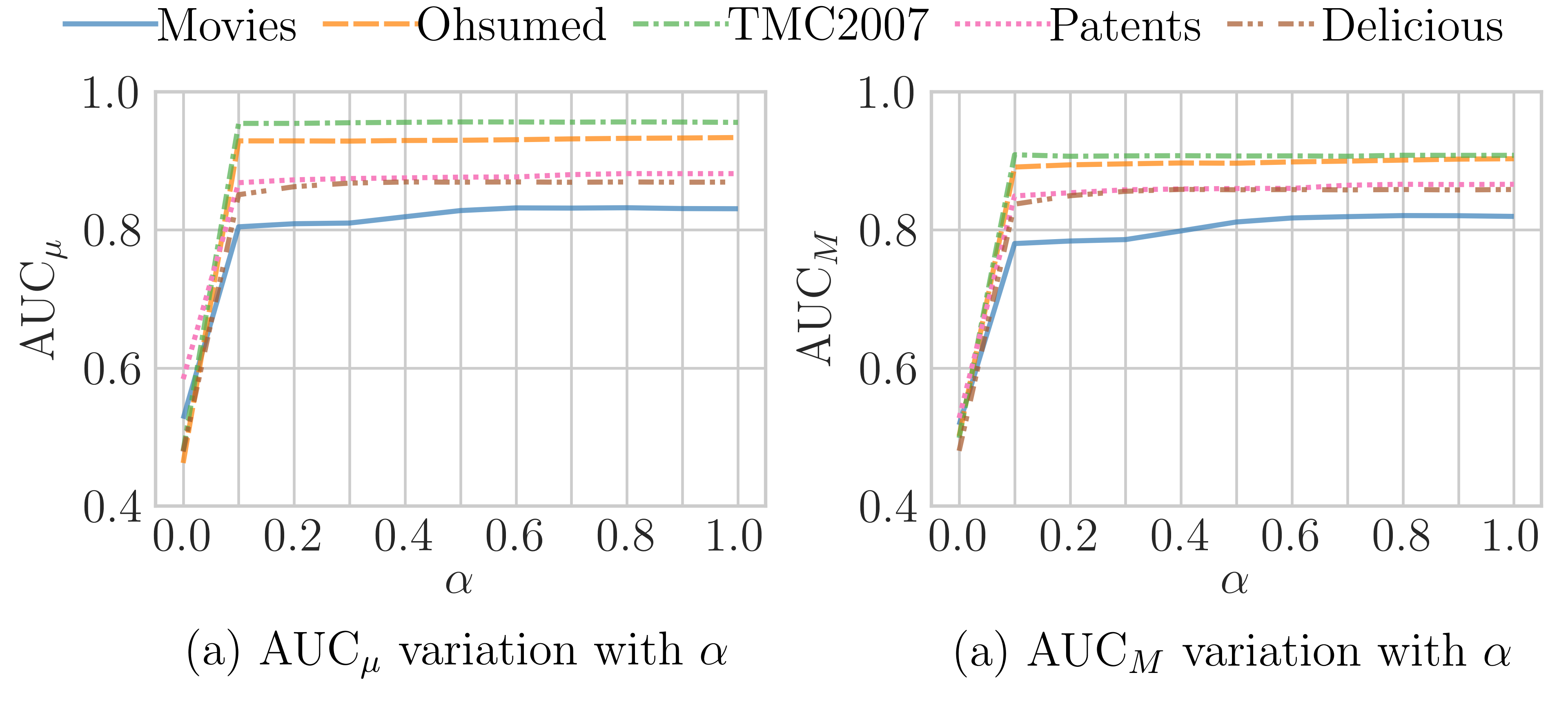}}
\caption{Sub-figures (a) and (b) shows the change in $\auroc_{\mu}$ and $\auroc_{M}$, with change in $\alpha$, respectively. }
\label{fig:alpha_auc}
\end{figure}

Figures~\ref{fig:alpha_auc}(a) and ~\ref{fig:alpha_auc}(b) show the change in performance on the $\auroc_{mu}$ and $\auroc_{M}$ metrics with change in $\alpha$, respectively. For both the metrics, the performance increases with an increase in $\alpha$, i.e., the performance on the $\auroc$ metrics is negatively impacted by the attention loss. As explained earlier, attention loss ignores the sentences that provide the negative signals for the classes to make its predictions, thus, adversely affects the multi-label classification performance.

The same behaviour is shown by the performance on the $\pppa$ metric, as depicted in Figure~\ref{fig:alpha_pppa}.

\section{Conclusion}\label{conclusion}
In this paper, we proposed \emph{Credit Attribution With Attention (CAWA)}, an end-to-end attention-based network to perform credit attribution on documents.
CAWA addresses the limitations of the prior approaches by (i) modeling the semantically similar classes by modeling each sentence as a distribution over the classes, instead of mapping to a single class; and (ii) leveraging a simple average pooling layer to constrain the neighboring sentences to have similar class distribution. A loss function is proposed to constrain that the attention is only focused on the classes present in the document. 
The experiments demonstrate the superior performance of CAWA over the competing approaches. 
Our work makes a step towards leveraging distant-supervision for credit attribution and envision that our work will serve as a motivation for other applications that rely on the labeled training data, which is expensive and time-consuming. As a future work, an interesting direction is to use distant supervision for image segmentation and annotation. Code and data accompanying with this paper are at https://github.com/gurdaspuriya/cawa.

\section*{Acknowledgment}

This work was supported in part by NSF (1447788, 1704074, 1757916, 1834251), Army
Research Office (W911NF1810344), Intel Corp, and the Digital Technology Center at the
University of Minnesota. Access to research and computing facilities was provided by
the Digital Technology Center and the Minnesota Supercomputing Institute.

\bibliographystyle{aaai}  
\fontsize{9.0pt}{10.0pt}
\bibliography{refs}
\end{document}